# Analyzing Polysemy Evolution Using Semantic Cells

Yukio Ohsawa   Dingming Xue   Kaira Sekiguchi

*The University of Tokyo, Tokyo, Japan  {ohsawa,kaira}@sys.t.u-tokyo.ac.jp, xdingming@outlook.com*

**Abstract.** The senses of words evolve. The sense of the same word may change from today to tomorrow, and multiple senses of the same word may be the result of the evolution of each other, that is, they may be parents and children. If we view word senses as an evolving ecosystem, the paradigm of learning the "correct" meaning, which does not move with the sense of a word, is no longer valid. This paper is a case study that shows that word polysemy is an evolutionary consequence of the modification of Semantic Cells, which has already been presented by the author, by introducing a small amount of diversity in its initial state. We show an example of analyzing the a set of short sentences. In particular, the analysis of a sentence sequence of 1000 sentences in some order of the four senses of the word "spring," collected using Chat GPT, shows that the word acquires the polysemy monotonically in the analysis when the senses are arranged in the order in which they have evolved. In other words, we present a method for analyzing the dynamism of a word's acquiring polysemy with evolution and, at the same time, a methodology for viewing polysemy from an evolutionary framework rather than a learning-based one.

# 1     Introduction

Word Sense Disambiguation (WSD) is considered a long-term task in Natural Language Processing (NLP). According to Navigli [Navigli et al 2013], the WSD task involves selecting the intended sense from a predefined set of senses for a word defined by a sense inventory. When the same word occurs in different senses, it can be represented as multiple tokens (spring_A, spring_B, etc.) or as multiple nodes in a network of words. Visualizing the network of word senses where multiple senses of a word appear as multiple nodes can also be positioned as a means of correctly understanding the sense of a word.

With the maturity of the distribution hypothesis, the literature posits that contextually similar words have similar semantics, making word embeddings different in different contexts and can address disambiguation to some extent [Wang et al 2020]. Since 2018, the mainstream approach has used embeddings i.e. context-embedded distributed representations, such as in BERT, where the polysemy of words and the diversity of expressions are pre-learned in the transformer model. LSTM-based ELMo [Peters et al 2018], transformer-based GPT, BERT [Devlin et al 2019] and other models have been established for generating dynamic, deeply contextualized embeddings that enable the grasping of the semantics of words with contextualized senses. In [Wiedemann et al 2019], they argued that a pre-trained BERT model could place distinct senses of a word corresponding to distinct places in the embedding space.

Here let us discuss - what is an "correct sense"? When machine learning methods are used to learn the sense of each word from existing textual information and store it in vector form, the larger volume of text, the better as long as there is one fixed correct sense. Rather than linking this to the classical law of large numbers, it is better to think of a reasoning process to logically reach the correct sense by clearing up misunderstandings by adding evidence, since some information is ambiguous or lacks evidence to identify the sense, and to directly obtain an understanding of the sense with high explainability [Ohsawa et al 2023]. This principle holds true even if there is more than one correct sense. For example, if a sentence in which a word appears can have multiple clusters consisting only of words around it, the word sense can be learned for each cluster [Neelakantan et al 2014, Amrami et al 2018]. For example, if the word "spring" means spring as a season, it is learned as a word that appears in a context that co-occurs with the surrounding words "beginning," "hanami," "sakura," etc. If it means "coil spring," it is learned as a word that appears in a context that co-occurs with "toy," "pen," "sports," etc. The multiple senses of a word can also be located by visualizing which word it is connected to in a network of words [Jauhar et al 2015].

However, an important perspective is missing from this discussion - word senses *evolve*. The word "spring" now has at least four senses: "spring (as water flow)," "spring (as a season)," "leap," and "spring (a coil which jumps)." It is thought that there was a process of semantic evolution in which the next sense was derived from the immediately preceding sense. As discussed below, "spring" is a word that has been variously derived and evolved from the verb meaning "to erupt or leap."

A similar phenomenon is vastly more prevalent today: the term Artificial Intelligence (AI) progressed in its early years to topics about robots that could play with building blocks, and then came to be interpreted as if it were synonymous with machine learning. Some now consider it synonymous with the Chat GPT. The evolution of the term AI may appear to be different from that of spring because it inherited the original sense of 'the ability of a machine to perform intelligent tasks like human" and only changed its way of embodiment, but in fact spring has also added senses depending on the necessary context in its history of expressing a bouncy divergent power. In fact, spring is also the same as AI in that it has added senses according to the necessary context in the history of expressing the bouncy divergent force.

Thus, words usually evolve in such a way that new senses are added while inheriting the original sense; however, senses that have not been used for a long time are forgotten and removed from the word's list of senses. For example, in spring, the original use of the word in the sense of "first appearance; beginning, birth, origin" has now largely disappeared. In other words, the polysemy of a word can be thought of as a process of birth and growth along an evolutionary sequence, while simultaneously eliminating the older senses. The senses currently used for a word may follow this process of evolution and selection, and those that have often been eliminated do not exist in a visible form.

Arranging the current uses of words in the order of their evolving senses can illustrate the process of the gradual emergence of senses. For example, a sentence using "spring" meaning to leap

> The frog springs from the lily pad.
> The rabbit springs out of the bushes.
> He springs forward to catch the ball.

In these three sentences a "spring" indicates that the animal (including human) moves at once to a high and wide place, i.e., a leap. Next, we read:

> They built a dam to control the flow of the spring.
> The spring was a lifeline for the nomadic tribe.
> The spring water was used for making herbal medicine.

This means a natural extension of the context from one in which only animals appear into a context that also includes the movement of water. Furthermore, we read:

> The toy's spring mechanism was fascinating.
> The sofa has a comfortable spring system.
> The watch's spring was wound too tightly.

These refer to objects that can leap away from a small restrained state. They are derived from animal leaps rather than from water which is not elastic. Thus far, we have a hypothesis that explains the process by which the three senses emerged. If these examples appeared much later, it would have been difficult to explain them in relation to examples we have seen in the past. In that case, one might think that there are only two senses of spring, the second and third, without being able to recall the first sense.

In this paper, the authors present an example of analyzing polysemy evolution by improving previously proposed Semantic Cells (SCs) [Ohsawa et al 2024]. In the above example, spring is considered to have evolved in the following order: "gush (or leap)," "spring as water flow" or "spring as a leaping coil," and "spring as a season." An experiment is presented to compare the differentiation process of senses by the crossover of chromosomes between cells, corresponding to words, in a sentence when the examples are arranged in a similar order and when they are arranged in a different order. If we can observe the gradual emergence of the four existing senses from the list of sentences in an evolutionary order and the disappearance of the previous senses in a different order due to contextual disorder, this would suggest that Semantic Cells are capable of correctly capturing the emergent dynamism of polysemy.

## 2 Semantic Cells

The aim of crossover between chromosomes is to differentiate the senses of words, that is, to simulate the emergence of polysemy. In this section, we explain the procedure of how SCs work referring to Figure 1.

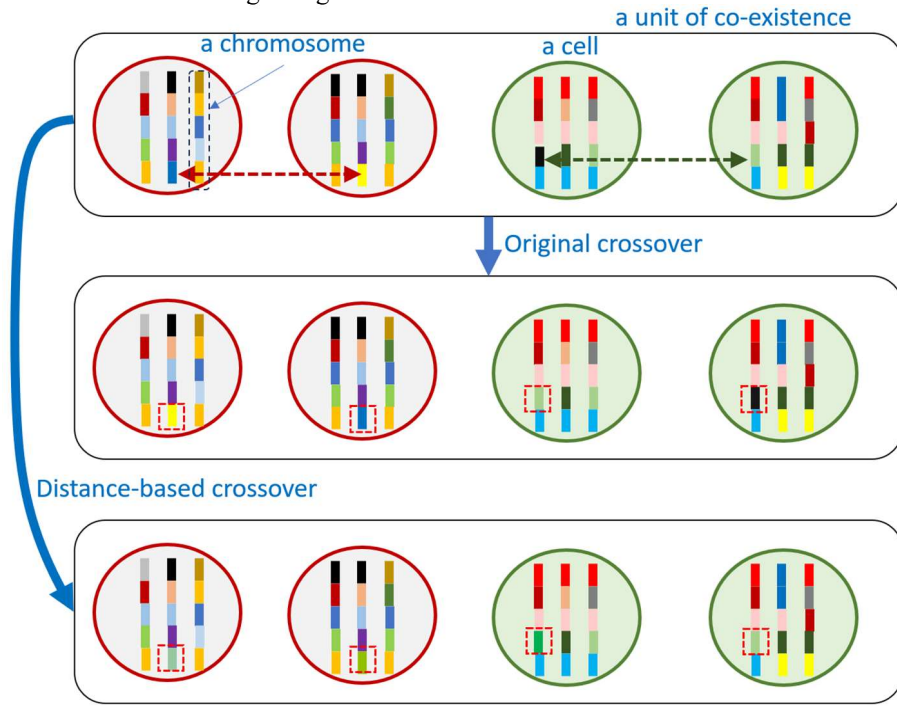

Fig. 1. The crossovers between chromosomes in cells, each of which corresponds to an item (word, item, event, etc.). Multiple chromosomes in a cell correspond to the multiple senses of an item, among those within close relationships (similarities) execute a crossover. The crossover is simply realized in this study by reducing the distance between two chromosomes by changing the values of elements of the vectors corresponding to each genetic element (Ohsawa et al 2024).

The process starts setting all chromosomes nearly equal (adding small values for initial differentiation of chromosomes in each cell as explained below) within each cell to the embeddings of the corresponding word, and a chromosome in each cell nearest to the average vector of all the chromosomes in a sentence increases the similarity to it in a step of crossover. The chromosomes obtained as a result of the crossover process represent the diversity within a cell. We presented SC for the first time in [Ohsawa et al 2024], to which here we add improvement with succeeding the text in order to clarify the improvement.

As shown in Figure 1, a Semantic Cell $C_w$ is assigned to item $w$ in itemset $W$. Here, an item refers to a word if the dataset is text, a commercial item if POS data, or an earthquake or its focal point (or the epicenter) if it is historical seismic data. Here, an item refers to a word if the dataset is text, a commercial item if POS data, an earthquake, or its focal point (or epicenter) if it is historical seismic data. $C_w$ includes $g$ chromosomes, which are vectors that represent the senses of an item. Here, a sense corresponds to the meaning if the item is a word, the customer's utility if the item is a commercial item sold in the market, or the land-crust movements if the item is an earthquake.

If item $w$ occurs with other items in the $j$-th unit of co-existence, for example, a sentence in text or a basket in the POS data of a supermarket, the chromosomes go through the process of crossover, which has been used in genetic algorithms [Goldberg and Lingle 1985, Goldberg 1989, Eiben and Smith 2015ab]. In contrast to existing genetic algorithms, where parts of a chromosome are exchanged with others, here the chromosomes for the crossover operation are seleted on the distance with the average chromosome, i.e. the average vector, in the same unit (the closer it is, the more likely it is). The number of crossovers equal to the number of crossovers, or the number of coexisting units in the dataset, is called one round and $R$ rounds are executed in the crossover process. The final vectors, that is, chromosomes, as a result of the crossover process, represent the diversity within a cell.

The evolution of semantic cells can be described as follows, where $d$ represents the dimension of each vector, which is the length of, that is, the number of genes in, each chromosome. $g$ is the number of chromosomes in each cell (i.e., item). $R$ is the number of rounds in which the crossover is iterated, which was set to one in this study. In each crossover operation, $\text{Ch}_S$ represents the vector representing the co-existence unit $S$. The crossover in Figure 1 is executed by reducing the distance between the chromosome selected from each cell and all other chromosomes in the belonging co-existence unit, where the chromosome is selected from a cell if it is in the shortest distance from $\text{Ch}_S$. Although $\text{Ch}_S$ can be computed as the vector of the co-existence unit using the method for doc2vec [Le and Mikolov 2014], here, we used the sheer average of all the vectors in all the cells in unit $S$ for simplicity. $\alpha$ Here represents the influence of a unit on the selected chromosome of an item during the crossover operation. $\alpha$ may be replaced $by\ \alpha/r^2$ where $r$ represents the distance above considering the analogy with gravity or Coulomb's force between two particles, we used only $\alpha$ in order to accelerate the crossover of items that have been located far in the distance-based crossover in Line 15.

In line 4, we applied word2vec to the Gensim library [Řehůřek 2022] to obtain the initial chromosomes, which are semantic vectors of words, where $d$ was set to 50 and $g$ to 5. These settings may be simpler than real natural chromosomes, and than the real natural language or standard values of $d$ because this is a preliminary experiment.

However, it has been studied that larger dimensions do not necessarily result in a more accurate analysis [Melamud et al 2016].

As an improvement to the previous work [Ohsawa et al 2024], as in line 6, the $d$ genes in each chromosome were divided into $g$ segments of $d/g$ genes and added $\epsilon$ (<<1: 0.01 here) to each of the $(di/g)$+1 th through the $d(i+1)/g$ th genes. Thus, each chromosome was prepared for the crossover with chromosomes in other words in multiple contexts, corresponding to different dimensions of the semantic vectors. This is an improvement for avoiding the repeated selection of the same chromosome in a cell as a candidate for the crossover with chromosomes in other cells or the average of all the vectors, i.e., chromosomes in all the words in a sentence.

1: **Input:** 1
2: Dataset $D$ including co-existence units $\{S|\ S \subset D\ \}$
3: Set of items $\{w|$ each item in unit $S$ including $|S|$ items$\}$
4: The $j$-$th$ chromosome $\mathrm{Ch}_{wj}$ , of item $w$ including $g$ chromosomes
5: Each gene: (1) $\mathrm{chr}_{wjk} = \mathrm{chr}_{wjk} + \delta \mid 0 < j \leq g,\ 0 < k \leq d, w \in W$
6:     where $\delta = \epsilon$ if $\mid (di/g\ ) < \mathrm{k} < 1 + d(i+1)/g$ , $else\ \epsilon/(g-1)$
7: **Output**
8: Evolved chromosomes $\mathrm{Ch}_{wj}$ for each $w \in W\ and\ j \in g$
9: **Crossover iteration.**
10: for each $r$ in $[1:R]$.
11:   for each $S \subset D$
12    $\mathrm{Ch}_S := average\ \big(\ \mathrm{Ch}_{wj}\big)_{\square} s.t. w \in S, j\ in\ [1:g]$
13:    for each $w \in S$
14:         $\mathrm{Ch}_{wj0} := the\ j_0$ th chromosome in $w$, of the least dist. in $w$ from $\mathrm{Ch}_S$
15:         $\mathrm{Ch}_{wj0} \leftarrow (1 - \alpha)\mathrm{Ch}_{wj0} + \alpha\ \mathrm{Ch}_S$
16:     end for
17:   end for
18: end for

## 3    Results

According to the section on "springs" in Etym online (https://www.ety-monline.com/word/spring), a natural spring, which refers to "water gushing to the surface of the earth or the flow of water from underground to the surface," was described in Old English (until the 12th century) as a "spring," "source," or "splash." However, the word spring has been rarely used alone. The word spring, sense "fountain," is an adaptation of the earlier verb spring, "to gush out," to water.

The word "erupt" may be considered similar to the verb "leap," as a phenomenon. The noun meaning "the act of jumping or leaping" began to be used in the late 14th century, and the name was applied in the early 15th century to an elastic metal coil that returns to its original shape after being pulled. However, the spring itself, which is not limited to metal, is older than the history of the English language, and as far as the author has been able to determine, the first time it was called a spring was not known.

The sense of spring as a season appeared in the 1540s as a contraction of spring of the year (1520s); "spring of day" as sunrise, "spring of month" as moonrise in the 14th century. It is certain that they appeared relatively late, considering that they are derived from the noun spring, which expresses "the act or time of springing or appearing, the first appearance, and the beginning, the origin."

Thus, "spring" is thought to have evolved in the following order:

(1) Verb "to erupt" or "to leap."

(2) "spring as fountain" or "spring as coil," a noun derived from (1)

(3) "spring as a season," a noun derived from "beginning" common to (1) and (2).

Therefore, we conducted an experiment to simulate and compare the process of differentiation of senses, or the emergence of polysemy, by the crossover of chromosomes among cells in a sentence. The examples are arranged in a similar order to history and in a different order for comparison. The results are shown in Figure 2.

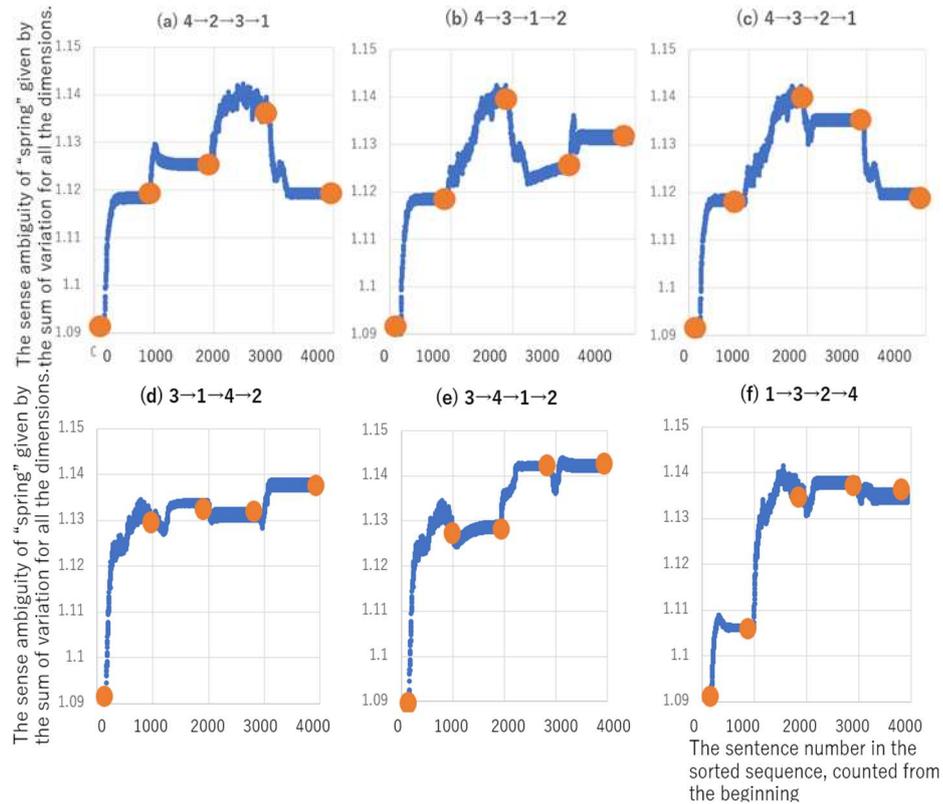

Figure 2. 1000 short sentences containing each sense of "spring," arranged in various semantic orders (4000 sentences in total), read and processed by the modified SC procedure from the beginning to the end. The four digits connected via arrows show the senses at the top (1: fountain, 2: a season, 3: leap, 4: leaping coil).

In Figure 2, The horizontal axis shows the number of sentences read by SC, and the vertical axis shows the acquired polysemy, given by the diversity of chromosomes, calculated as the sum of the variance of the $g$ values of genes for all the d dimensions.

Here, we do not present numerical evaluations of the results, such as accuracy or computation time, because all we know is that at least these four senses exist for "spring" and that they are presumed to have increased in the above order. Therefore, we sheerly expected that if the sentences in the experimental sequence were in the above order, the four senses would gradually manifest themselves, and the polysemy would increase monotonically. On the other hand, if the sequence deviates from the above order, the context may change abruptly and the number of senses may decrease instead of monotonically increasing by the four senses because the previous senses are forgotten on the way. The aim of this experiment was to confirm these phenomena. Therefore, with the acquired polysemy (the diversity of chromosomes, calculated as the sum of the variance of the $g$ values of genes for all the $d$ dimensions) on the vertical axis and the number of sentences read by the SCs on the horizontal axis, we can see whether these phenomena are observed. In Figure 2, (a), (b), and (c) show the results of having the SCs read a randomly ordered sequence of the four senses of spring, 1000 sentences presented by ChatGPT4o for each sense, and (d) and (e) are the two variations that follow the historical order of the appearance of senses we mentioned in (1), (2), and (3) above. In the last, (f) follows the order of sense evolution according to ChatGPT4o, i.e., the author asked "in which order did the four senses (spring as a season, spring as a coil, spring as a fountain, and spring as to leap) evolve for spring?"

## 4    Discussions

From the above results, it can be seen that (a), (b), and (c) in Figure 2 are the cases in which polysemy decreases on the way, especially with a significant decrease in (b) even before the last. In contrast, in (d) and (e), polysemy increases nearly monotonically. In (f), there is a slight drop at the end, but the increase fits the monotonic uptrend. It is possible that ChatGPT4o used for generating the sentence-sequence for (f) does not correctly show the history of the four senses but rather a sequence of continuous changes in the context of the four senses. From this point of view, the results of (f) are not in line with the purpose of the evolutionary order that is the subject of this paper, but are not far removed from the purpose of the evolutionary sequence that this paper is trying to deal with. In this case, too, the acquisition of polysemy is seen to be progressing at a nearly monotonic uptrend compared to (a), (b), and (c).

There are at least three reasons for the importance of the evolution of word sense. First, since the sense of a word is always evolving, the vector set (i.e., set of chromosomes) of the known word at the latest time can be obtained by such a model as SC where the vectors evolve by time. Applying SCs to WSD, we can expect to improve the accuracy even if new senses of the word are emerging nowadays. This has a significant effect on technical terms in the latest technological fields or words used by people in the young generation, where senses evolve unexpectedly quickly. The sense of a word may change even within a domain according to the contextual shifts although the change may be moderate e.g. "sister " may not suddenly turn to mean a "nun" within a story but change from one's sister living in Tokyo into a member of a new family in London. Grasping such a change in sense due to a contextual shift is also an expected contribution of SC.

Second, since all senses that have evolved in the past are the origins of the sense of the word, it is possible to develop a method that captures the sense of a word not as a point corresponding to a semantic vector, but as a line, which is the trajectory of points which appear during the history of the word. In other words, we can study and acquire the history structure of a word by choosing the order of example sentences using the word in respective senses in which the curve such as in Figure 2 comes to show a monotonic uptrend. Such a study has been studied by collecting historical use cases and investigating the year and the context of the appearance of each case, e.g., spring appeared in Old English to mean "source" or "splash," which takes a huge volume of efforts, cost, and time.

Third, by applying a model that simulates the evolution of a word with a known order of lexical evolution, as in the experimental case above, and see if the results match the known order, we can evaluate whether the model can correctly capture polysemy. If results such as those in Figure 2 are verified for words in rich data on text, then SC is superior as a polysemy detection method.

All in all, SC can be regarded as a method to support sensemaking in data [Dervin 1992], which is a process for collecting, representing, and organizing information for decision-making in real tasks [Russell et al. 1993, Weick 1993]. The original concepts of sensemaking has been extended to various studies chance discovery which has been initiated in 2000 and lasing today ([Ohsawa and Mcburney 2003] etc). Sensemaking is supposed to be triggered by recognizing the inadequacy of the current understanding of events to explain their meaning [Klein et al. 2006b]. Awareness of the ambiguity of the sense of an event, item, or word triggers sensemaking. The SC model is expected to aid both this awareness and the choice and explanation of the most suitable sense that fits the situation.

## 5    Conclusions

The analysis of four sentence sequences of 1000 sentences, listed in various orders for the four senses of the word "spring" collected using Chat GPT 4o, showed that the word acquires polysemy monotonically by SCs when the senses are arranged in the order in which they have evolved historically. Thus, we presented a method for analyzing the dynamism of a word's acquiring polysemy with evolution and, at the same time, a methodology for viewing polysemy from an evolutionary framework rather than a learning-based one. That is, the SC model does not aim to learn "correct" senses of words but to discover "emerging" senses.

On the other hand, the senses of items in each application domain can be put into vectors; thus, semantic vectors can be diverted to various domains. That is, the senses of items (or events) in each domain can be put into vectors; thus, semantic vectors can be diverted to various domains. Similar diversions have been found in various domains:

e.g. co-occurrence based analyses of earthquakes [Ohsawa 2002, Fukui et al 2014] and vectors for n-grams in biological sequences (e.g., DNA, RNA, and proteins) have been proposed [Asgari and Mohammad 2015], and commercial items in the market have also been vectorized [Barkan and Koenigstein 2016, Cao et al 2022]. In the extension of this study, it is expected that a basic SC algorithm can be diverted to the vectorization of various items to understand the senses (or the ways to use) of items or events in the real world. We have already shown some results for assessing the local risks of earthquakes, as in the original version of SC [Ohsawa et al 2024]. In the extension of this study, it is expected that SCs can be used to externalize the history of evolution in various domains in nature and human society, which emerge via dynamic interactions among items.